\documentclass[letterpaper, 10 pt, conference]{ieeeconf}  

\IEEEoverridecommandlockouts                              

\usepackage{xcolor}
\usepackage{graphicx}
\usepackage{amsmath}
\usepackage{amssymb}
\usepackage{flushend} 
\usepackage[hidelinks]{hyperref}
\usepackage{threeparttable}
\let\labelindent\relax
\usepackage{enumitem}
\usepackage{float}
\usepackage{stfloats}

\title{\LARGE \bf
The Robot in the Room: Influence of Robot Facial Expressions and Gaze on Human-Human-Robot Collaboration
}

\author{Di Fu*, Fares Abawi, and Stefan Wermter
\thanks{The authors are with the Knowledge Technology Group, Department of Informatics, University of Hamburg,
22527 Hamburg, Germany
{\tt\small \{di.fu; fares.abawi; stefan.wermter\}@uni-hamburg.de}}
\thanks{*Corresponding author 
{\tt\small di.fu@uni-hamburg.de}}
}

\begin{document}

\bstctlcite{IEEEexample:BSTcontrol}
\maketitle
\thispagestyle{empty}
\pagestyle{empty}

\begin{abstract}

Robot facial expressions and gaze are important factors for enhancing human-robot interaction (HRI), but their effects on human collaboration and perception are not well understood, for instance, in collaborative game scenarios. In this study, we designed a collaborative triadic HRI game scenario, where two participants worked together to insert objects into a shape sorter. One participant assumed the role of a guide. The guide instructed the other participant, who played the role of an actor, by placing occluded objects into the sorter. A humanoid robot issued instructions, observed the interaction, and displayed social cues to elicit changes in the two participants' behavior. We measured human collaboration as a function of task completion time and the participants' perceptions of the robot by rating its behavior as intelligent or random. Participants also evaluated the robot by filling out the Godspeed questionnaire. We found that human collaboration was higher when the robot displayed a happy facial expression at the beginning of the game compared to a neutral facial expression. We also found that participants perceived the robot as more intelligent when it displayed a positive facial expression at the end of the game. The robot's behavior was also perceived as intelligent when directing its gaze toward the guide at the beginning of the interaction, not the actor. These findings provide insights into how robot facial expressions and gaze influence human behavior and perception in collaboration.
\end{abstract}

\section{INTRODUCTION}

\textcolor{black}{Collaboration is a fundamental aspect of human social behavior, which plays a crucial role in achieving common goals and solving problems~\cite{ohtsuki2006simple}. However, collaboration can be challenging, and conflicts may arise. One potential mitigation to this issue is the integration of humanoid robots into human collaboration settings. Humanoid robots can potentially assist humans in enhancing their collaboration skills~\cite{strohkorb2016improving}. \textcolor{black}{For example,} robots can be useful in \textcolor{black}{engaging children with autism spectrum disorders} or in reducing conflicts during collaboration~\cite{scassellati2018improving}. However, it is essential to acknowledge that while robots can be helpful in such interactions, they may only have limited influence due to the complexity of social dynamics. Therefore, understanding human collaboration behavior and how robots could impact it is critical to successfully integrating humanoid robots into social settings.}

\begin{figure}[!hbtp]
\centering\includegraphics[trim={0.01cm 0 0 0},clip,width=0.48\textwidth]{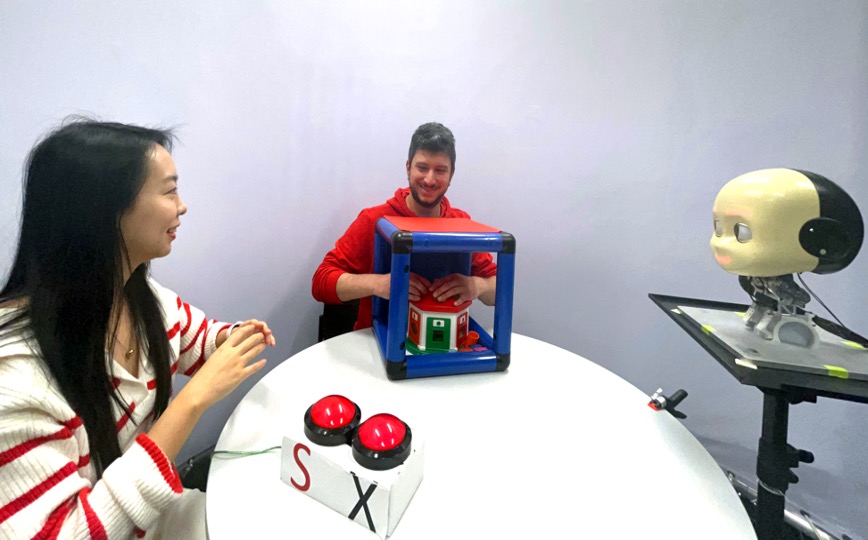}
  \caption{Experimental setup. Two participants adopt the roles of a guide (left) and an actor (center) while the iCub robot (right) observes their gameplay. The actor inserts an object into the sorter following the guide's instructions.}\label{fig:intro_demo}
\end{figure}

\textcolor{black}{Social cues influence human behaviors and responses in interactional situations, including human-human collaboration. Non-verbal communication through facial expressions, body language, and prosody assist humans in interpreting the emotions and intentions of others~\cite{fu2021trained}. These social cues provide vital information that allows us to understand social norms, establish trust, and form positive relationships. Humans are more likely to collaborate and work together \textcolor{black}{towards} common goals when they receive positive social cues. Negative social cues, on the other hand, can lead to mistrust, conflict, and a breakdown in collaboration~\cite{tomasello2013origins}. As a result, social cues are considered crucial in shaping human behavior and collaboration~\cite{tanis2003social}.}

\textcolor{black}{Eye gaze and facial expressions are particularly important social cues in HRI and collaboration. Humanoid robots are designed to simulate human behaviors and express emotions through movement and facial expressions. When robots display positive social cues like maintaining eye contact, nodding, and smiling, they can establish a connection with humans and gain their trust~\cite{babel2021small}. This can lead to more effective human-robot collaboration in a variety of tasks, including manufacturing, healthcare, and education~\cite{terziouglu2020designing}. Furthermore, social robots can facilitate human-human or human-robot collaboration by signaling them and acting as mediators~\cite{andriella2021have}. Overall, social cues in HRI can enhance communication, making their integration into robots a positive design choice.}

\textcolor{black}{Although previous studies have shown robot gaze and facial expressions to have an impact on HRI, limited \textcolor{black}{research has} explored the impact of humanoid robots in \textcolor{black}{triadic --- human-human-robot ---} collaboration scenarios. Moreover, \textcolor{black}{only a }few studies investigated the interaction effect between gaze and facial expressions from robots on human collaboration behaviors. Thus, in this study, we created a collaborative game between two human participants with the aim of inserting objects into a shape sorter~\cite{fu2022judging}. One participant served as a guide, giving instructions to the other participant, who acted as an actor by placing occluded objects in the sorter. A humanoid robot was incorporated into the setup, displaying facial expressions while directing its gaze toward either the actor or the guide.}

Our goal is to determine how much the presence of a robot influences human behavior and perception, both explicitly through measures like button-pushing and questionnaires and implicitly by measuring their game completion time.  It is important to note that we do not aim to study whether a robot's presence in such situations influences the interaction. Instead, we assume it to be present and investigate its impact on human interaction. 

\textcolor{black}{In the current work, we investigate how a humanoid robot's non-verbal social cues, specifically facial expressions and gaze communication, affect \textcolor{black}{triadic} collaboration. We propose two main research questions (RQ):}

\begin{description}
    \item \textbf{RQ1:} \textcolor{black}{How can a robot's facial expressions and gaze communication impact triadic collaboration? }
    \item \textbf{RQ2:} How do humans perceive the intelligence of a robot during \textcolor{black}{triadic} collaboration? Is it consistent with humans' general impressions of the robot?
\end{description}

\noindent \textcolor{black}{from which we derive the following hypotheses:}

\begin{description}
    \item \textbf{H1:} The robot's positive facial expressions will improve human-human collaboration performance compared with neutral facial expressions. A mutual gaze between the guide and the actor could impact the performance of the task differently. They may have an interaction effect on human collaboration between facial expressions and gaze.
    \item \textbf{H2:} The robot's positive facial expressions will make individuals perceive the robot as more intelligent than the neutral and negative facial expressions. A mutual gaze between the guide and the actor could elicit participants to have different impressions of the robot. There could be an interaction effect on human perception of the robot between \textcolor{black}{facial expressions and gaze}. 
\end{description}
 
  \textcolor{black}{This research aims to explore humanoid robots' potential as collaborators in human-human teams and their ability to communicate effectively through non-verbal social cues.}

\section{RELATED WORK}

\textcolor{black}{Robot gaze significantly influences on human decision and perception. Kompatsiari et al.~\cite{kompatsiari2017importance} studied the effects of mutual and non-mutual robot gaze. Their findings revealed that participants attribute greater engagement and human-like traits to a robot in establishing eye contact. Another study has shown mutual gaze between robots and humans to influence the latter's decision-making time~\cite{belkaid2021mutual}: Participants were slower at making decisions when the iCub robot established eye contact with them. Neural activity in the brain evoked by the robot's gaze draws similarity to gaze influences observed during social interactions with other humans, indicating that robot gaze has a similar effect as human gaze~\cite{belkaid2021mutual}. Moreover, eye contact with robots elicits physiological changes associated with positive affect and higher attention allocation~\cite{kiilavuori2021making}.}
\textcolor{black}{Gillet et al.~\cite{gillet2021robot} investigated how a social robot could use adaptive gaze behavior to balance the participation of a native speaker and a second language learner in a game. \textcolor{black}{These} results show that the robot's gaze could influence interaction among players leading to an even contribution in participation between them.}

\textcolor{black}{Robot emotional cues, whether through speech, gestures, facial expressions, or other indications of affect, alter humans' perception of the robotic agent~\cite{saunderson2019robots}. Their mental states are also influenced through emotion contagion~\cite{neumann2000mood}. Reyes et al.~\cite{reyes2016positive} studied how a human-like robot's sad facial expressions on failing to complete the task affected human-robot collaboration. The task was to place ten objects in a container by collaborating with a robot. The authors~\cite{reyes2016positive} found that the robot's sadness signaled a need for human help and improved task performance. In a follow-up work~\cite{reyes2019robotics}, the authors suggested that negative facial expressions signaling failure attract humans' attention and lead them to collaborate better.}

\textcolor{black}{Unlike previous studies that focused on direct HRI, our main objective is to evaluate the influence of non-verbal cues, namely, robot facial expressions and gaze behavior, on \textcolor{black}{triadic HRI}. We \textcolor{black}{hypothesize} that this question is important to understand how a robot can facilitate social dynamics among humans without interfering with their verbal communication.  A robot, as an observer, can also elicit different responses from humans depending on how they perceive the robot's human likeness, intelligence, and intentionality. Therefore, studying how a robot can use non-verbal cues to modulate human-human interaction helps us \textcolor{black}{to} design social robots that can improve their collaboration.}

\section{METHODOLOGY}

To investigate our research questions and examine our hypotheses, we made two participants play a collaboration game while the iCub robot joined them as an observer in the experiment. We measured participants' completion time of the game and recorded their perceptions of the robot's intelligence during the game. Participants were also asked to \textcolor{black}{fill in} the Godspeed questionnaire after the game to report on their impression of the iCub robot.

\subsection{Participants} 
50 participants (female = 13, male = 37, non-binary = 0, prefer not to say = 0) took part in this experiment. Participants were between 21 to 55 years of
age, with a mean age of 29.02 ± 5.60 years. All participants
reported no history of neurological conditions (seizures,
epilepsy, stroke, etc.) and had either normal or corrected-to-normal vision and hearing. This study was conducted
following the principles expressed in the Declaration of
Helsinki. Each participant signed a consent form approved by the Ethics Committee of the Department of Informatics, University of Hamburg.

\subsection{Task and Procedures}

\textcolor{black}{In our study, we randomly matched participants in pairs. Each pair played multiple rounds of a \textcolor{black}{triadic} collaboration game while the iCub robot observed their interaction. Additionally, the iCub robot adopted the role of an instructor, requesting participants to place a particular object in its corresponding hole on a shape sorter. One of the two participants played the role of an actor and was capable of manipulating the objects and the shape sorter, which were obscured from their view. The other participant, having an unobstructed view of the objects and shape sorter, guided the actor in placing the right object into its designated hole. Guidance was restricted to non-physical contact, conveyed mainly through verbal instructions and physical gestures.}

\begin{figure*}[!hbtp]
\centering\includegraphics[trim={0 0 0 0},clip,width=0.90\textwidth]{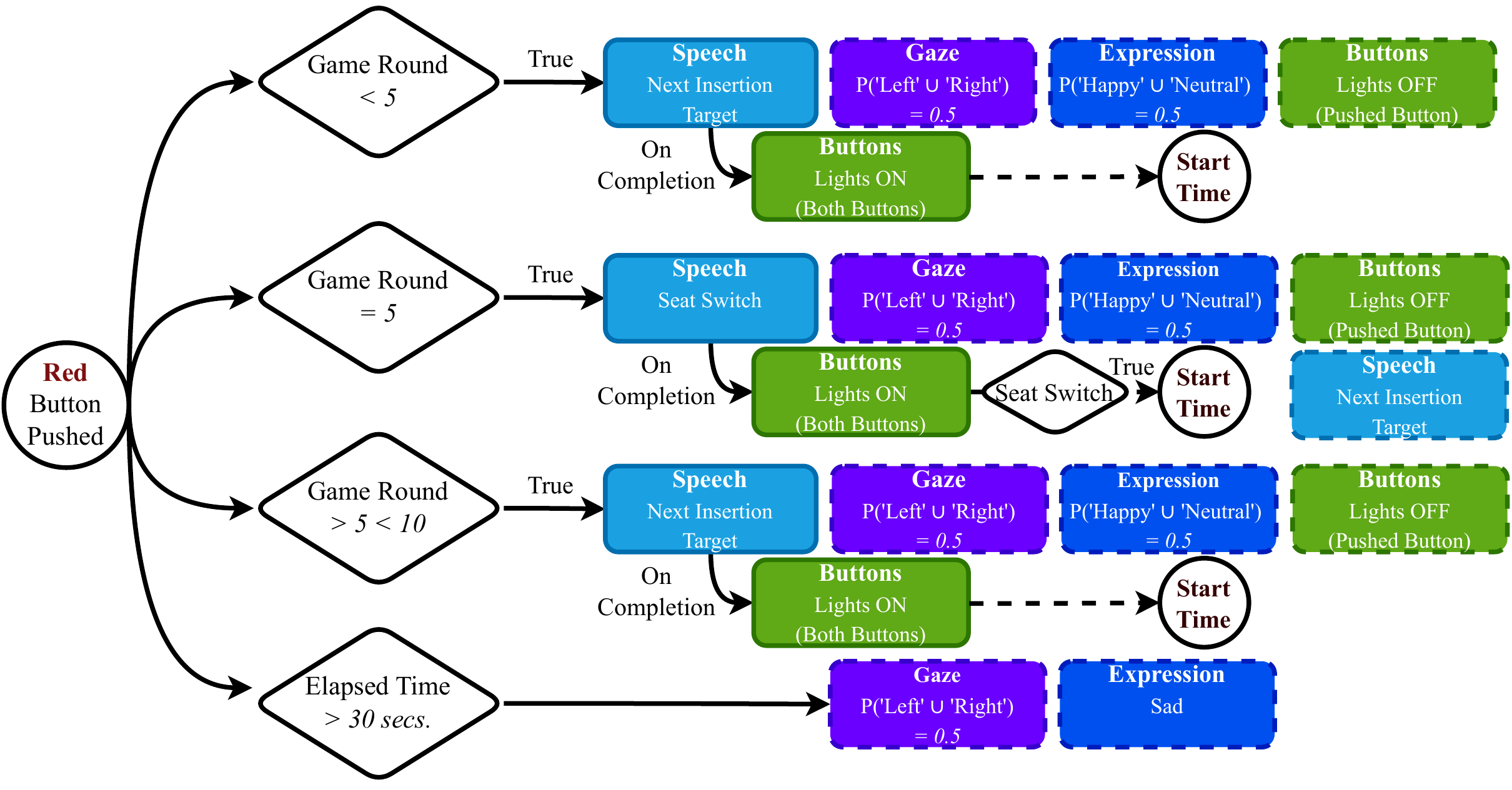}
  \caption{Game flow under different conditions after pressing either rating button. Speech is uttered initially on fulfilling a condition. Once the utterance is completed, follow-up actions are executed (e.g., switching on the button lights). Action blocks with dashed borders are executed in parallel after follow-up actions are completed, or previous conditions are fulfilled. }\label{fig:task_overview}
\end{figure*}

\textcolor{black}{The participants played a total of 10 rounds. Before each round started, the iCub robot displayed an initial facial expression, which could be neutral or happy. During the game, they were tasked with the successful insertion of an object into the shape sorter. After 5 rounds of gameplay, participants changed seats, consequently switching their roles as actors and guides. After each round, the iCub robot displayed its final facial expressions. The participants were asked to guess the intention behind the robot's facial expressions. In doing so, the participants would distribute their attention between the task at hand and the iCub robot. Each round completion time was recorded as the collaboration time for each pair. On round completion, participants were requested to rate the iCub robot as either intelligent or random according to its gaze behavior and facial cues during the round. The robot rating was performed by the participant assuming the role of a guide. Using two separate buttons to categorize the iCub robot as either intelligent or random, we acquired their responses following each round, along with their round completion time. If the participants finished one round within 30 seconds, the iCub robot kept its final facial expression as the initial one. If the participants failed to complete the task within 30 seconds, the iCub robot displayed a sad final facial expression and shifted its gaze either toward the actor or the guide. After completing all 10 rounds, the participants filled in the Godspeed questionnaire~\cite{bartneck2008measuring} to report their impression of the iCub robot based on its appearance and behavior. The overall flow of the game is illustrated in~\autoref{fig:task_overview}.}

\subsection{Experimental Setup} 

\textcolor{black}{The experimental setup consisted of a round table where the iCub head was placed, and two human participants were seated. The distance between the iCub's head and each participant was approximately 140 cm. A shape sorter with 12 holes, each with a color --- 6 colors in total --- corresponding to an animal and a basic-shaped object, was placed on the table. The sorter and objects were occluded from the actors' view by opaque surfaces covering the sides of a plastic wireframe. The guide had a clear view of the shape sorter in order to guide the actor in inserting an object specified by the iCub robot.}

\begin{figure}[!hbtp]
\centering\includegraphics[trim={0.01cm 0 0 0},clip,width=0.485\textwidth]{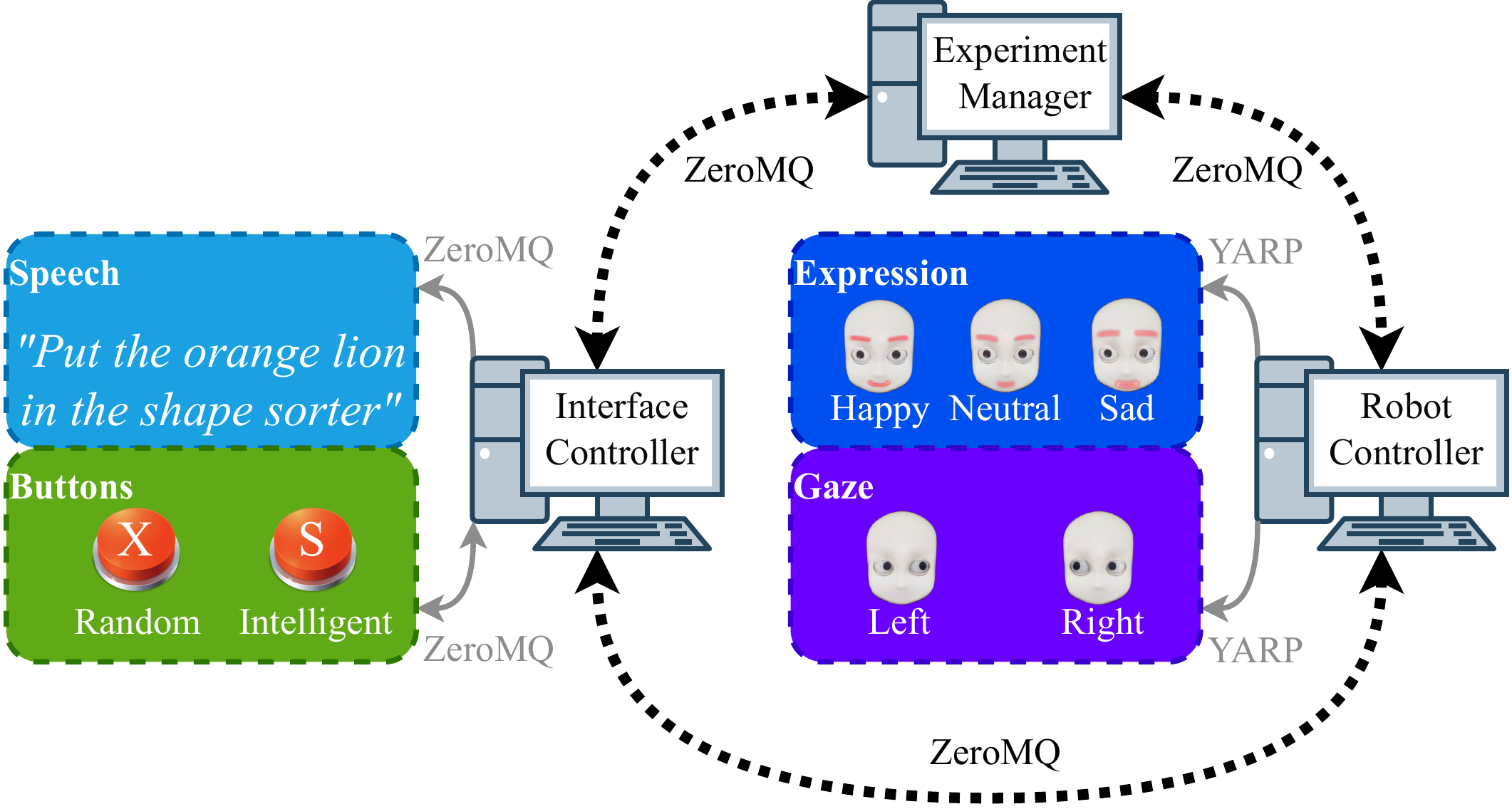}
  \caption{Experiment manager defines the game flow and communicates with controllers (represented by dotted arrows). Controllers connect unidirectionally to devices except for buttons since they register the user input and transmit it to the controller, requiring bidirectional communication.}\label{fig:communication_overview}
\end{figure}

\textcolor{black}{After each trial, the guide rated the iCub robot's intelligence by pressing one of two labeled red buttons with lights. The `X' labeled button indicated the iCub robot's observed behavior followed an unspecified pattern that did not correlate with the participants' actions. The `S' labeled button signified an intelligent pattern of the iCub robot's behavior, which is associated with the participants' gameplay. The button lights signaled the ongoing running of the experiment round. Pressing either button momentarily switched it off until instructions for the next round were verbally delivered through a loudspeaker, situated behind the iCub robot. Instructions were simply structured phrases to convey the target for each round, e.g., `Put the orange lion in the shape sorter'. These instructions were uttered using Amazon Polly speech synthesis, spoken with a child voice labeled as `Justin' to match the iCub robot's appearance.}

\textcolor{black}{We define the game flow as the full experimental pipeline, beginning with the iCub robot introducing the game to participants, followed by providing instructions on which objects to insert, and eventually thanking the participants for taking part in the experiment. Further processes involved in the game flow include providing instructions on switching seats and filling in the questionnaire, keeping track of the participants' game completion time, performing gaze movements, and displaying facial expressions. The game flow involves three computers with different roles as depicted in~\autoref{fig:communication_overview}:}
\begin{enumerate}
    \item \textcolor{black}{The \textbf{Experiment Manager (EM)} runs the main script, which coordinates the tasks of controllers that interact with external devices and sensors. The EM receives feedback from the controllers and delegates actions to them, such as moving the iCub's head in either direction, changing the iCub robot's facial expression, or uttering instructions. It communicates over the ZeroMQ~\cite{zeromq2013hintjens} middleware using Wrapyfi~\cite{abawi2023wrapyfi}, a Python wrapper with multi-middleware support for exchanging native Python objects, tensors, and arrays.}
    \item \textcolor{black}{The \textbf{Interface Controller (IC)} awaits button presses by the participants who had to rate the iCub robot's behavior as intelligent or random. It also controls the embedded button lights and sends audio signals to the speech interface via ZeroMQ. The buttons are connected to an Arduino AT-Mega 2560 microcontroller that communicates with the IC over USB serial. The IC also uses Wrapyfi for communicating over ZeroMQ.}
    \item \textcolor{black}{The \textbf{Robot Controller (RC)} sends control signals to the iCub robot to make it gaze toward the guide or the actor based on predefined estimated positions. It also sends emotion templates to the robot through the emotion interface. Since the iCub robot runs YARP~\cite{yarp2006metta}, we utilize both YARP and ZeroMQ on the RC to communicate with the iCub robot and the EM, respectively.}
\end{enumerate}

\subsection{Data Analyses}

\textcolor{black}{Completion time (CT) and rating of the robot are measurements of participants' game performance and perception of the robot, respectively. We conducted a two-factor repeated measures ANOVA with facial expressions (neutral vs. happy) and gaze direction (actor vs. guide) on the game completion time to examine the impact of the robot's \textbf{initial} facial expressions and gaze on \textcolor{black}{triadic} collaboration. The \textbf{final} facial expressions and the gaze from the iCub robot were displayed after the participant finished each round of the game. Thus, their completion time was not affected by the robot's behavior after the game. Analyzing the impact of the robot's final facial expressions and gaze direction on the game performance is, therefore, not required.}

\textcolor{black}{To investigate how the robot's initial facial expression and gaze direction influence participants' perception of the robot's intelligence, a two-factor repeated measures ANOVA with facial expressions (neutral vs. happy) and gaze direction (actor vs. guide) was performed on participants' ratings. We encoded participants' `intelligent' rating of the robot with a value of `1', and `random' with a value of `0'. Thus, the higher ratings the robot got, the more intelligent participants perceived it.}

\textcolor{black}{To measure the impact of the robot's final expression and gaze on participants' ratings of the robot, one paired $t$-test was performed between sad and happy expressions. Another paired $t$-test was performed between the ratings of the actors and the guides. We did not analyze the interaction effect between the final expressions and gaze. This is due to participants observing more sad expressions than happy and neutral expressions since their completion time was usually longer than 30 seconds. Under the sad expression condition, gaze direction was balanced. However, under the happy and neutral expression conditions, gaze direction was not balanced, resulting in a majority of participants experiencing only a subset of the condition combinations.}

\textcolor{black}{Additionally, we also investigated whether there would be any differences between the first and last 5 rounds of participants' game performances and robot ratings by using independent $t$-tests, given that participants switched roles after 5 rounds. Eventually, we analyzed the correlation between completion time, robot rating, and five sub-dimensions of the Godspeed Questionnaire to study the relationship between participants' general impression of the robot and their perception of it during the game. All post hoc tests in the current study used Bonferroni correction.}

\section{RESULTS}

\begin{figure*}
    \centering
    \begin{minipage}{0.5\textwidth}
        \centering
        \includegraphics[width=0.95\linewidth,trim={0.4cm 0.2cm 0.4cm 0.2cm},clip]{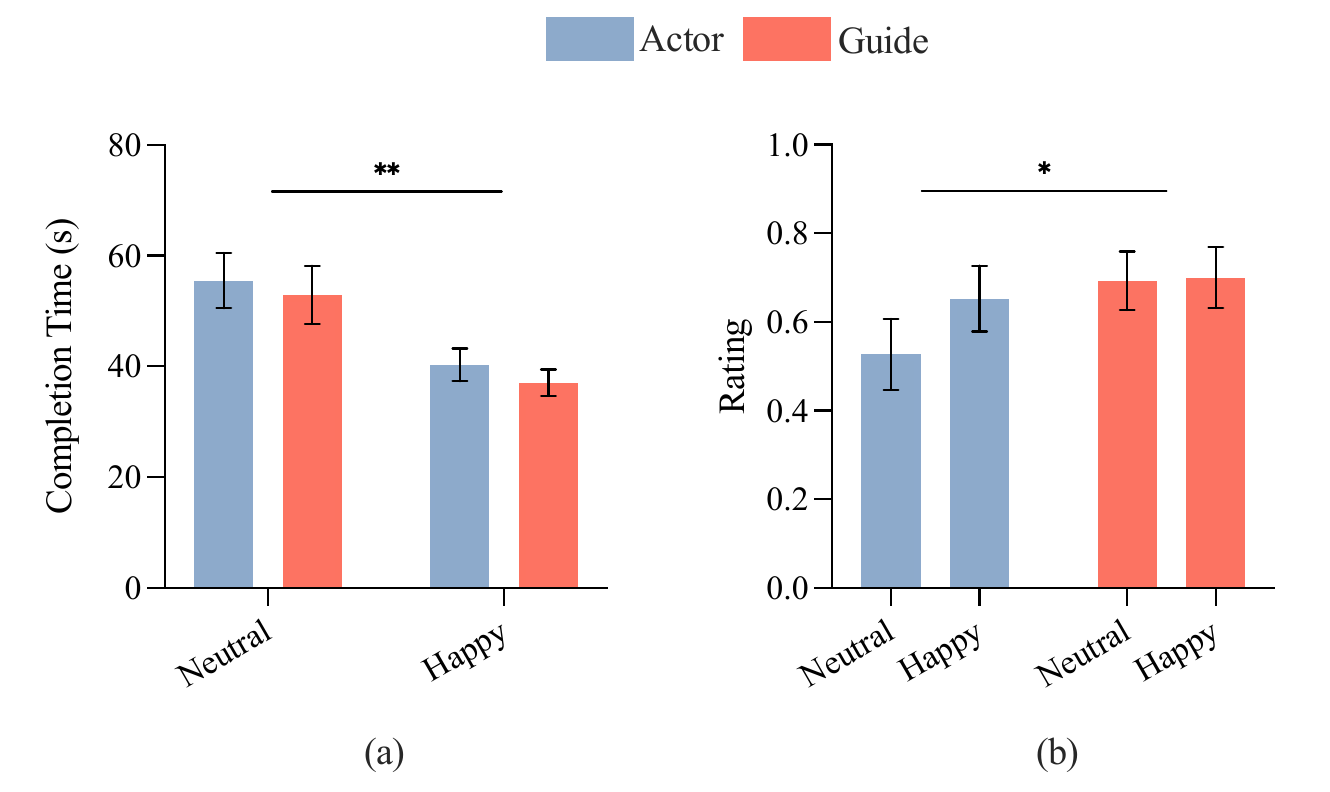}
    \end{minipage}\hfill
    \begin{minipage}{0.5\textwidth}
        \centering
        \includegraphics[width=0.95\linewidth,trim={0.4cm 0.2cm 0.4cm 0.2cm},clip]{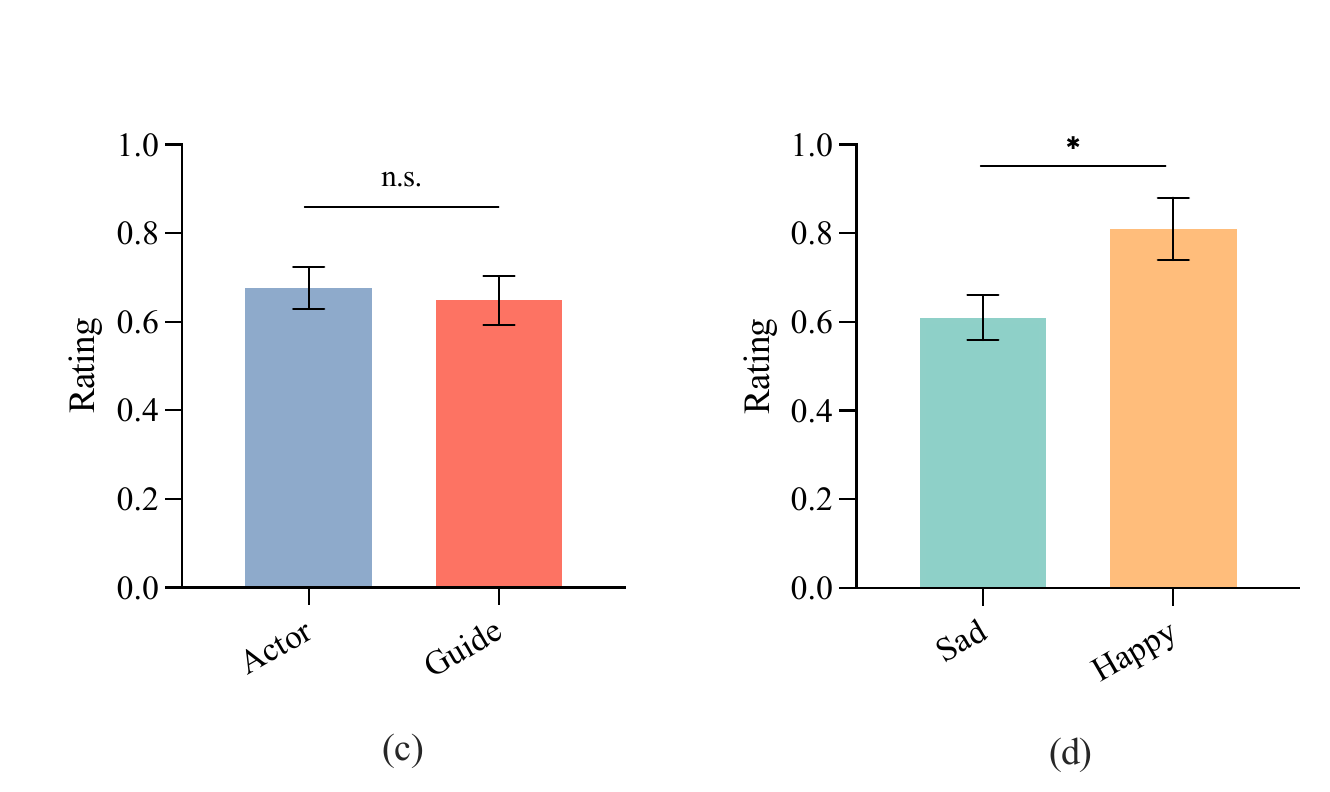}
    \end{minipage}
    \begin{tablenotes}
      \footnotesize
      \item  \hspace{1ex} CT: Round completion time; Robot Rating: participants' evaluation of the robot after each game round. 
      \item \hspace{1ex} $*$ denotes $.01 < p < .05$, $*\!*$ $.001 < p < .01$, and \textit{n.s.} denotes no significance.
    \end{tablenotes}
    \caption{Participants' completion time of the game \textbf{(a)} and their rating of the robot's intelligence \textbf{(b)} under its different initial facial expression and gaze direction conditions; Their rating of the robot's intelligence given its different final gaze directions \textbf{(c)} and facial expressions \textbf{(d)}.}\label{fig:main_results}
\end{figure*}

In this section, we report our results on participants' completion time and their perceptions of the robot during and after the game.
\subsection{Initial Facial Expressions and Gaze on Collaborative Game Performances}
\textcolor{black}{To evaluate the impact of the initial facial expressions and gaze on the collaborative game performances, a repeated measures ANOVA with a Greenhouse-Geisser correction was applied. Results displayed in \autoref{fig:main_results}a showed that the main effect of facial expressions was significant. The participants' RT differs significantly between different facial expression conditions, $F \left( 1, 23 \right) = 15.73, p < .01, \eta_{p}^{2} = .40$. Post hoc tests show that the participants finished the game significantly faster under the happy condition ($\text{mean} \, \pm \, \text{SE} = 38.66 \pm 2.04 \, \text{ms}$) than the neutral condition ($\text{mean} \, \pm \, \text{SE} = 54.22 \pm 4.15 \, \text{ms}$). However, the main effect of the initial gaze direction was not significant. There was no significant difference in participants' game performances, whether or not the robot's initial gaze was toward the actor or the guide, $F \left( 1, 23 \right) = .93, p = .35, \eta_{p}^{2} = .04$. There was no significant interaction effect between the initial facial expressions and the initial gaze, $F \left( 1, 23 \right) = .01, p = .94, \eta_{p}^{2} = .00$.}

\subsection{Initial Facial Expressions and Gaze on Rating the Robot}
\textcolor{black}{To evaluate the impact of the initial facial expressions and gaze on rating the iCub robot, a repeated measures ANOVA with a Greenhouse-Geisser correction was applied. Results presented in \autoref{fig:main_results}b showed that the main effect of facial expressions was not significant. There was no significant difference in participants' ratings of the robot between neutral ($\text{mean} \, \pm \, \text{SE} = .61 \pm .06 \, \text{ms}$) and happy conditions ($\text{mean} \, \pm \, \text{SE} = .68 \pm .06 \, \text{ms}$), $F \left( 1, 23 \right) = .79, p = .38, \eta_{p}^{2} = .03$. However, the main effect of the initial gaze direction was significant, $F \left( 1, 23 \right) = 4.94, p < .05, \eta_{p}^{2} = .17$. Post hoc tests show that the participants perceived the robot significantly more intelligent when the robot initially looked at the guide ($\text{mean} \, \pm \, \text{SE} = .70 \pm .05$) than when looking at the actor ($\text{mean} \, \pm \, \text{SE} = .59 \pm .06$). There was no significant interaction effect between the initial facial expression and the initial gaze on participants' ratings of the robot, $F \left( 1, 23 \right) = 1.18, p = .29, \eta_{p}^{2} = .05$.}

\subsection{Final Facial Expressions and Gaze on Rating the Robot}
\textcolor{black}{Paired-samples $t$-tests were conducted to study the influence of the final emotion on rating the robot. Results in \autoref{fig:main_results}d showed that participants rated the iCub robot significantly more intelligent when the robot displayed happiness ($\text{mean} \, \pm \, \text{SE} = .81 \pm .07$) than sadness ($\text{mean} \, \pm \, \text{SE} = .61 \pm .05$), $t \left( 24 \right) = -2.46$, $p < .05$.}

\textcolor{black}{Paired-samples $t$-tests were performed to investigate how the final gaze direction impacted the robot's rating. No significant difference was found between the two conditions (actor: $\text{mean} \, \pm \, \text{SE} = .68 \pm .05$, guide: $\text{mean} \, \pm \, \text{SE} = .65 \pm .05$, $t \left( 24 \right) = .52$, $p =.61$, see \autoref{fig:main_results}c).}

\subsection{Learning Effects in the Collaborative Game}

\begin{figure}[!hbtp]
\centering\includegraphics[trim={0.9cm 0.35cm 0.2cm 3.2cm},clip,width=0.450\textwidth]{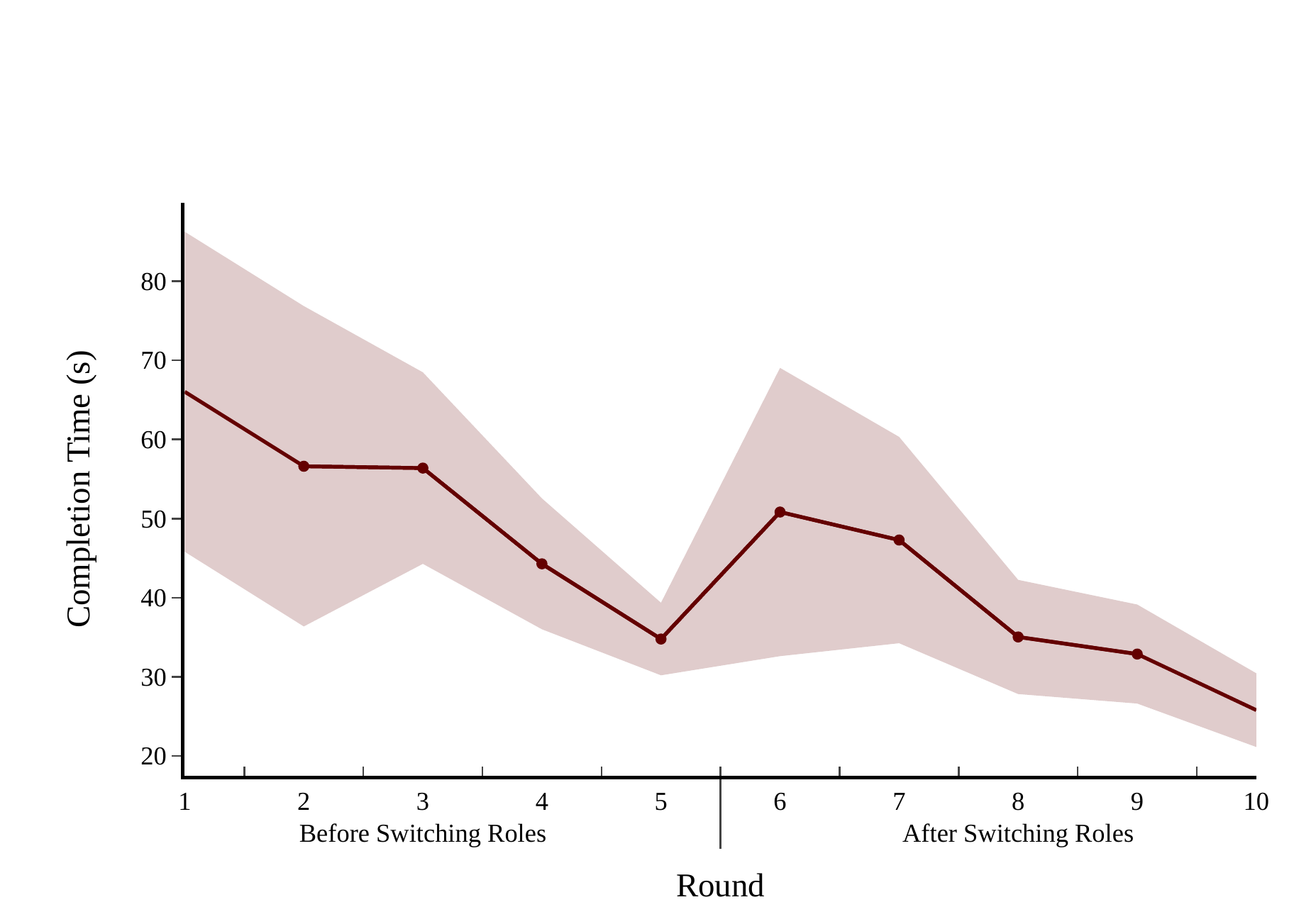}
  \caption{The mean completion time and the standard error per game round decline as participants gain collaboration and gameplay experience.}\label{fig:completion_rounds}
\end{figure}

\begin{figure}[!hbtp]
\centering\includegraphics[trim={0.9cm 0.35cm 0.2cm 3.2cm},clip,width=0.450\textwidth]{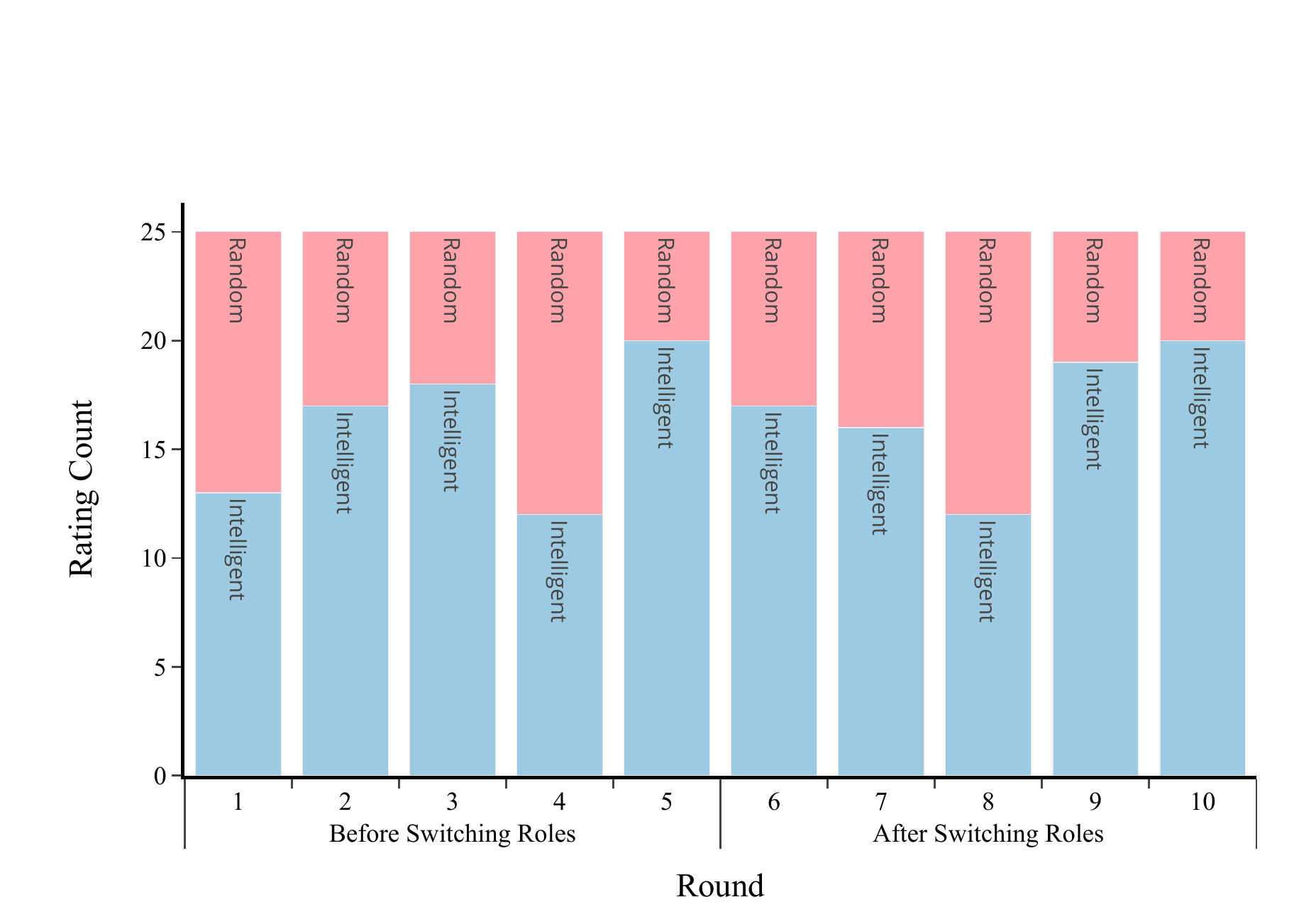}
  \caption{Robot rated as intelligent more often than random. Intelligence rating increases from 16 before switching roles to 16.8 times after the switch.}\label{fig:robot_rating}
\end{figure}

\textcolor{black}{\autoref{fig:completion_rounds} shows that there was a reduction in completion time for the first 5 rounds of the game. After switching roles, participants' completion time in the last 5 rounds also decreased, indicating that learning effects persisted throughout the game. Furthermore, we conducted independent paired-samples $t$-tests showing that participants took significantly less time completing the last 5 rounds ($\text{mean} \, \pm \, \text{SE} = 38.38 \pm 2.12$ s) than the first 5 rounds ($\text{mean} \, \pm \, \text{SE} = 51.62 \pm 3.32$ s), $t \left( 24 \right) = 3.36$, $p < .01$. These findings imply that repeated exposure to the collaborative game resulted in improved performance, emphasizing the importance of experience and practice in improving collaborative skills. However, robot ratings did not follow a consistent trend. Independent paired-samples $t$-tests performed on robot ratings indicated no significant difference between the first 5 rounds ($\text{mean} \, \pm \, \text{SE} = .64 \pm .05$) and the last 5 rounds ($\text{mean} \, \pm \, \text{SE} = .67 \pm .06$), $t \left( 24 \right) = -3.81$, $p = .71$. These results suggest that participants' perception of the robot did not change with more practice of the game (rating counts after different rounds are shown in \autoref{fig:robot_rating}). }

\subsection{Godspeed Questionnaire}

\begin{table*}[!hbtp]
\centering
\caption{Means, Standard Deviations, and Correlation Matrix of the Measurements}
\label{tab:my-table}
\begin{tabular}{lllllllll}
\hline
 & Mean ± SD & CT (s) & Robot Rating & Anthropomorphism & Animacy & Likeability & Intelligence & Safety \\ \hline
CT (s)         & 45.10 ± 15.33 & \phantom{---}1    &                        &        &        &        &                         &   \\
Robot Rating & \phantom{---}.66 ± .29     & \phantom{-}.04  & \phantom{--------}1                      &        &        &        &                         &   \\
Anthropomorphism        & \phantom{--}2.48 ± .86    & -.01 & \phantom{------}.24                    & \phantom{-----------}1      &        &        &                         &   \\
Animacy                 & \phantom{--}2.74 ± .83    & \phantom{-}.06  & \phantom{------}.22                    & \phantom{---------}.82*** & \phantom{-----}1      &        &                         &   \\
Likeability             & \phantom{--}3.51 ± .86    & \phantom{-}.03  & \phantom{------}.27 & \phantom{---------}.61*** & \phantom{---}.66*** & \phantom{------}1      &                         &   \\
Intelligence  & \phantom{--}3.10 ±\phantom{-}.84     & -.10 & \phantom{------}.41**                  & \phantom{---------}.74*** & \phantom{---}.68*** & \phantom{-----}.71*** & \phantom{-------}1                       &   \\
Safety                  & \phantom{--}3.66 ± .72    & -.10 & \phantom{-----}-.08                   & \phantom{---------}.36**  & \phantom{---}.48*** & \phantom{-----}.56*** & \phantom{------}.27 & \phantom{---}1 \\ \hline
\end{tabular}
\begin{tablenotes}
  \footnotesize
  \item  \hspace{1ex} CT: Round completion time; Robot Rating: participants' evaluation of the robot after each game round. 
  \item \hspace{1ex} $*$ denotes $.01 < p < .05$, $*\!*$ $.001 < p < .01$, $*\!*\!*$ $ p < .001$, and \textit{n.s.} denotes no significance.
\end{tablenotes}
\end{table*}

\textcolor{black}{Means and standard deviations for completion time, robot rating, five sub-dimensions (Anthropomorphism, Animacy, Likeability, Perceived Intelligence, Safety) of the Godspeed questionnaire, as well as the correlation coefficients between them, are displayed in table 1. The rating of the robot during the game was positively correlated to Perceived Intelligence ($r = .41, p < .01$). The Completion time was not significantly correlated with any other measurements ($ps > .05$). Additionally, a weak positive correlation was observed between robot rating and Likeability ($r = .27, p = .063$). Within the sub-dimensions of the Godspeed questionnaire, only the association between Perceived Intelligent and Safety was marginally significant ($r = .27, p = .056$). Associations between other dimensions reached significance ($ps < .05$).}

\section{DISCUSSION}

\textcolor{black}{Our study showed that a robot displaying a positive (happy) facial expression on initiating interaction improves collaboration between humans: participants completed the task within a shorter period of time --- less than 30 seconds --- when the iCub robot appeared happy. We hypothesize that emotional contagion plays a role in altering the participants' emotions. The iCub robot's expression of happiness reflected positively on the participants' mood, resulting in them being more productive and collaborative. This hypothesis is supported by studies examining the relationship between emotional states and productivity~\cite{oswald2015happiness,world2020future}, indicating that happy individuals tend to have better performance.}

\textcolor{black}{Participants completing the task within 30 seconds also rated the iCub robot as more intelligent, even though the robot followed the same strategy in every interaction. One influencing factor could be that the iCub robot displayed a negative (sad) facial expression when the participants took longer than 30 seconds to complete the task. A robot that displays a happy facial expression may be perceived as more friendly, trustworthy, and competent than a robot that displays a negative emotion~\cite{calvo2020effects}. However, we were unable to examine the effect of deferred facial expressions --- shown after 30 seconds --- due to the limited sample size. Examining whether a display of negative emotions has an effect on the robot's intelligence rating would only be possible if we were to vary the facial expressions when participants took longer than 30 seconds to complete a round. Given the infrequent occurrence of the event, the two conditions would not result in sufficient samples for a statically sound comparison.} 

\textcolor{black}{On establishing mutual gaze with the guide, the iCub was regarded as more intelligent than when looking at the actor. Given that the guide rates the robot, we compared their ratings under the condition of mutual gaze --- the iCub robot looking at the guide --- and looking elsewhere. Our results align with previous findings, indicating that mutual gaze caused participants to perceive a robot as more engaged, human-like, and attentive, eliciting them to attribute higher intelligence to it~\cite{kompatsiari2017importance,belkaid2021mutual}}.
 
\textcolor{black}{Several limitations in the current study could be addressed in future research. First, the gaze directions for the final facial expressions should be balanced to study their interaction effect on rating the robot. Second, more measurements could be conducted on participants' personality traits and their trust in the robot. This could lead to a deeper understanding of the current results. Finally, involving more emotions and increasing human-human interaction rounds could yield more nuanced findings. Addressing these limitations would lead to \textcolor{black}{an even} more comprehensive understanding of the impact of non-verbal social cues on \textcolor{black}{triadic} collaboration.}

\textcolor{black}{Our findings sparked several directions for future research. First, it would be valuable to investigate the interaction between robots' verbal and non-verbal social cues in human-robot or human-human collaboration. Investigating how the use of various emotional expressions in addition to verbal communication affects collaboration and productivity~\cite{van2022social} would also be an important path to explore. Additionally, in order to create effective human-robot communication, robots need to be able to understand natural language and respond accordingly. One promising approach is the use of large language models, such as InstructGPT~\cite{ouyang2022training} and LaMDA~\cite{thoppilan2022lamda}, which have the ability to generate human-like responses to user input. Furthermore, there is a need for participant diversity in future studies, such as including children or individuals with autism, to improve the way they interact with people and potentially inform the development of socially assistive robots for these populations.}

\section{CONCLUSIONS}
\textcolor{black}{We investigated the role of non-verbal social cues by a humanoid robot in \textcolor{black}{triadic} collaboration. We also evaluated humans' perception of the robot given such cues. Our findings revealed that positive robot facial expressions are crucial in enhancing \textcolor{black}{triadic} collaboration, whereas appropriate robot gaze, when combined with positive expression feedback, can boost the perception of the robot's intelligence. These results underscore the necessity of designing social robots that are capable of displaying non-verbal social cues to promote successful human-human and human-robot collaboration. Further research is needed to explore the underlying mechanisms of these effects and to develop more sophisticated and \textcolor{black}{natural} social robots that can effectively engage humans in collaborative settings.}





\section*{ACKNOWLEDGMENT}

The authors gratefully acknowledge partial support from the German Research Foundation DFG under project CML~(TRR~169).


\bibliographystyle{IEEEtran}
\bibliography{root}

\end{document}